\documentclass[runningheads]{llncs}

\usepackage[mobile]{eccv}

\usepackage{eccvabbrv}
\usepackage{multirow}

\usepackage{graphicx}
\usepackage{booktabs}

\usepackage[accsupp]{axessibility}  

\usepackage[dvipsnames]{xcolor}
\usepackage{lipsum}
\usepackage[absolute,overlay]{textpos}
\usepackage{graphicx}

\newcommand{\tocite}[1]{\textbf{\color{red}[TOCITE]}}

\newcommand\blfootnote[1]{%
  \begingroup
  \renewcommand\thefootnote{}\footnote{#1}%
  \addtocounter{footnote}{-1}%
  \endgroup
}
\newcommand{\urlNewWindow}[1]{\href[pdfnewwindow=true]{#1}{\nolinkurl{#1}}}
\newcommand{\PAR}[1]{\noindent{\textit{\textbf{#1~}}}}

\usepackage[pagebackref,breaklinks,colorlinks,citecolor=eccvblue]{hyperref}

\usepackage{orcidlink}

\begin{document}

\title{SAM-guided Graph Cut \texorpdfstring{\\}~for 3D Instance Segmentation}

\titlerunning{SAM-guided Graph Cut for 3D Instance Segmentation}

\author{
    Haoyu Guo$^{1*}$
    \quad He Zhu$^{2*}$
    \quad Sida Peng$^{1}$
    \quad Yuang Wang$^{1}$\\
    \quad Yujun Shen$^{3}$
    \quad Ruizhen Hu$^{4\dagger}$
    \quad Xiaowei Zhou$^{1\dagger}$ \\
}

\authorrunning{Guo. et al.}

\institute{$^{1}$Zhejiang University $^{2}$Beijing Normal Univeristy $^{3}$Ant Group $^{4}$Shenzhen Univeristy}

\maketitle

\begin{abstract}

\blfootnote{$^{*}$ Equal contribution}
\blfootnote{$^{\dagger}$ Corresponding authors}
This paper addresses the challenge of 3D instance segmentation by simultaneously leveraging 3D geometric and multi-view image information. Many previous works have applied deep learning techniques to 3D point clouds for instance segmentation. However, these methods often failed to generalize to various types of scenes due to the scarcity and low-diversity of labeled 3D point cloud data. Some recent works have attempted to lift 2D instance segmentations to 3D within a bottom-up framework.
The inconsistency in 2D instance segmentations among views can substantially degrade the performance of 3D segmentation.
In this work, we introduce a novel 3D-to-2D query framework to effectively exploit 2D segmentation models for 3D instance segmentation.
Specifically, we pre-segment the scene into several superpoints in 3D, and formulate the task into a graph cut problem.
The superpoint graph is constructed based on 2D segmentation models, enabling great segmentation performance on various types of scenes. We employ a GNN to further improve the robustness, which can be trained using pseudo 3D labels generated from 2D segmentation models.
Experimental results on the ScanNet200, ScanNet++ and KITTI-360 datasets demonstrate that our method achieves state-of-the-art segmentation performance.
Our project page is available at \urlNewWindow{https://zju3dv.github.io/sam_graph}.
\end{abstract}
\section{Introduction}

\begin{figure}[t]
\centering
\includegraphics[width=\linewidth]{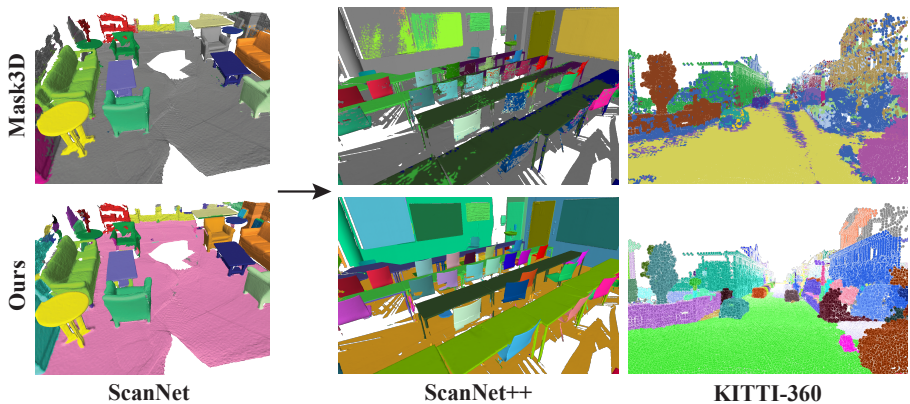}
\caption{
Thanks to the inductive bias introduced by SAM-annotated superpoint graph, our method achieves good segmentation performance and generalization capabilities. After training solely on ScanNet200, our model can effectively generalize to data collected with different devices (ScanNet++) and even to entirely different types of scenes (KITTI-360).
}
\label{fig:teaser}
\end{figure}

Instance segmentation of 3D scenes is a cornerstone of many applications, such as augmented and virtual reality, robot navigation, and autonomous driving.
One typical pipeline for 3D scene segmentation is using deep neural networks\cite{pointnet,pointnet++,pointcnn,kpconv} to process the point cloud of the target scene to predict segmentation results.
These methods\cite{mtml,bottomup1,bottomup4,topdown2,voting-based2,3d-mpa,pointgroup,softgroup,mask3d} generally require annotated point clouds for training.
However, annotating point clouds is costly, and thus there is a lack of datasets with the large scale and diversity similar to 2D image datasets. 
As a result, these methods are often limited to specific types of scenes and struggle to generalize to in-the-wild scenes.

Compared to point clouds, the acquisition and annotation of images are much less costly.
In recent years, with the emergence of large-scale labeled 2D datasets and improvement in model architecture and capacity, state-of-the-art 2D instance segmentation models \cite{sam,sam-hq,mask2former,cropformer} with strong generalization capabilities have been developed. 
Consequently, a natural approach of 3D segmentation is to lift multi-view 2D segmentations to 3D, leveraging them to achieve segmentation in arbitrary 3D scenes.

Recently, some 2D-to-3D lifting methods \cite{objnerf,objsdf,objsdf++,vmap,panoptic,contrastive,sam3d} have used a bottom-up framework, which first runs 2D segmentation on each view to obtain several masks, and then attempts to establish correspondences among masks across different views.
These masks are then merged in 3D to obtain the 3D segmentation results.
However, such a bottom-up approach has a significant issue: 2D segmentation masks from different views may be inconsistent.
For example, some instances may be segmented in some views but missing in others, thus severely degrading the performance of 3D segmentation.

In this paper, we propose a novel 3D segmentation approach based on a 3D-to-2D query framework, which effectively utilizes 2D segmentation models.
Unlike previous methods that generate multiple masks from multi-view images first, our approach begins with a pre-segmentation of the 3D scene into several superpoints.
Then, we construct a superpoint graph of the target scene and transform the problem into graph cut.
The edge weights of the graph are obtained by projecting graph nodes onto multiple views, using the prompt mechanism of SAM~\cite{sam} to predict multi-view masks, and calculating the intersection of corresponding masks.
The node features are obtained by aggregating multi-view image features.
Finally, we use a graph neural network to refine the edge affinity for the graph partition \cite{union-find}.
The SAM-based graph enables great segmentation performance on various types of scenes.
In addition, we develop a scheme to generate pseudo 3D labels from a 2D segmentation network and design a training strategy to effectively leverages these pseudo labels, allowing us to train our model without any manual 3D annotations.

We conduct experiments on the ScanNet200, ScanNet++ and KITTI-360 datasets. The experimental results indicate that, with the guidance of SAM in the construction of the graph, our method achieves good segmentation results on various types of scenes. Moreover, the GNN module in our method has good generalization capabilities, which is able to generalize well to other datasets with significant differences after trained on one dataset with pseudo-labels.

In summary, our contributions are as follows:
\begin{itemize}
    \item We propose a novel 3D-to-2D-query framework that leverages SAM to construct node features and edge weights of the superpoint graph, enabling great segmentation performance on various types of scenes.
    \item We employ GNN to improve the robustness and develop a scheme to generate pseudo 3D labels from a 2D segmentation network, enabling our model to be trained without any manual 3D annotations.
    \item We demonstrate state-of-the-art performance on ScanNet200 \cite{scannet200}, ScanNet++ \cite{scannet++} and KITTI-360 \cite{kitti360} datasets.
\end{itemize}
\section{Related work}

\PAR{3D scene segmentation.} The goal of 3D scene segmentation is to group the scene point cloud into semantically meaningful regions or distinct objects. Previous works leverage large-scale 3D labeled datasets to accomplish this objective in a supervised manner. They first train a neural network to extract per-point features, and then assign one predicted label to each point based on the extracted features. \cite{scnn,fcn,3dconv,semlabel,segcloud,pointnet++,pointcnn,spidercnn,kpconv} achieve semantic segmentation on point cloud, and \cite{mtml,bottomup1,bottomup4,topdown2,voting-based2,3d-mpa,pointgroup,softgroup} further distinguish between different objects with the same semantics, thus getting 3D instance segmentation results.
Recently, Mask3D\cite{mask3d} leverage Transformer\cite{transformer} to construct the segmentation network, attaining superior instance segmentation quality on 3D point clouds. 
3D-SIS\cite{topdown1} performs 3D instance segmentation on RGB-D scan data, fusing image features extracted from 2D convolution networks with 3D scan geometry features, allowing accurate inference for object bounding boxes, class labels, and instance masks.

Some works exploit 2D large vision-language models to achieve open-vocabulary segmentation in 3D space. OpenScene\cite{openscene} back-project per-pixel image features extracted by large vision language models from multi-view posed images to form a feature point cloud endowed with open-vocabularity abilities for various down-stream scene understanding tasks. PLA\cite{PLA} constructs multi-scale 3D-text pairs and uses contrastive learning to enable the model to learn language-aware embeddings for 3D semantic and instance segmentation.

Partitioning 3D point cloud into a collection of small, geometrically homogeneous regions, coined superpoints, can yield a decent rough prediction and effectively simplify the process of scene segmentation.
\cite{sp-affinity} and \cite{sp-graphcut} propose to represent each 3D scene by constructing a superpoint graph, where superpoints serve as graph nodes, and then based on this graph, 3D instance segmentation is performed by learning inter-superpoint affinity and clustering superpoint nodes into 3D objects.
\cite{robert2023efficient} propose an efficient semantic segmentation for large-scale scenes by partitioning point clouds into a hierarchical superpoint structure, and \cite{robert2024scalable} further extend it for panoptic segmentation by scalable graph clustering.
 
\PAR{2D-to-3D lifting.} Due to the lower cost of acquiring and annotating 2D images, the scale and diversity of 2D annotated datasets \cite{imagenet,mscoco,genome,cityscapes,pascal,lvis}, are much larger compared to 3D datasets, facilitating the emergence of many highly effective 2D segmentation methods \cite{sam,sam-hq,cropformer,mask2former} in recent years, making the use of 2D segmentation for 3D tasks a new and promising approach.
Semantic-NeRF \cite{semantic-nerf}, based on the NeRF framework, utilizes outputs from a 2D semantic segmentation network at each view to train a 3D semantic field. Since it can fuse information from multiple viewpoints, this method is robust to inaccuracies and noise in individual view segmentations, yielding better 3D semantic segmentation results.
However, extending this approach directly to instance segmentation is challenging, which is more complex since instance IDs in multi-view image segmentation results could inconsistent, making it necessary to design an effective label matching mechanisms in order to lift multi-view 2D instance segmentation results to 3D space.

To address this issue, \cite{panoptic} solves a linear assignment for instance identifiers across views with machine generated semantic and instance labels as supervision, \cite{contrastive} proposes a scalable slow-fast clustering objective function to fuse 2D predictions into a unified 3D scene segmentation results represented with a neural field.
SAD \cite{sad} uses SAM to segment both images and depth maps, combining the advantages of both for improved results.
SAM3D \cite{sam3d} proposes a method for point cloud fusion. It segments each frame and gradually merges the segmentation results of all frames together. For scenes with known geometry, it can achieve segmentation very efficiently.
SAI3D \cite{yin2023sai3d} partitions a 3D scene into geometric primitives, which are then progressively merged into 3D instance segmentations that are consistent with the multi-view SAM masks.
MaskClustering \cite{yan2024maskclustering} proposes a novel metric called view consensus to better exploit multi-view observation for 3D instance segmentation.
Some works integrate SAM \cite{sam} into the Neural Radiance Fields (NeRF) \cite{nerf} framework for 3D segmentation. For example, \cite{chen2023interactive} merges SAM features into Instant NGP \cite{instant-ngp} in 3D, allowing users to segment an object from 3D space through multiple clicks. OR-NeRF \cite{or-nerf} enables users to segment an object by clicking and then remove it from the scene.

Additionally, some methods combine segmentation and reconstruction, enabling separate reconstruction of each object. \cite{objnerf} propose to decompose a scene by learning an object-compositional neural radiance field, with each standalone object separated from the scene and encoded with a learnable object activation codes, allowing more flexible downstream applications. To cope with the ambiguity of conventional volume rendering pipelines, \cite{objsdf,objsdf++} further utilizes the Signed Distance Function (SDF) representation to exert explicit surface constraint.\cite{vmap} represents each object in the scene with a small MLP and builds an object-level dense SLAM that detects objects on-the-fly and dynamically adds them to its map.

\section{Method}

\begin{figure*}[t]
\centering
\includegraphics[width=\linewidth]{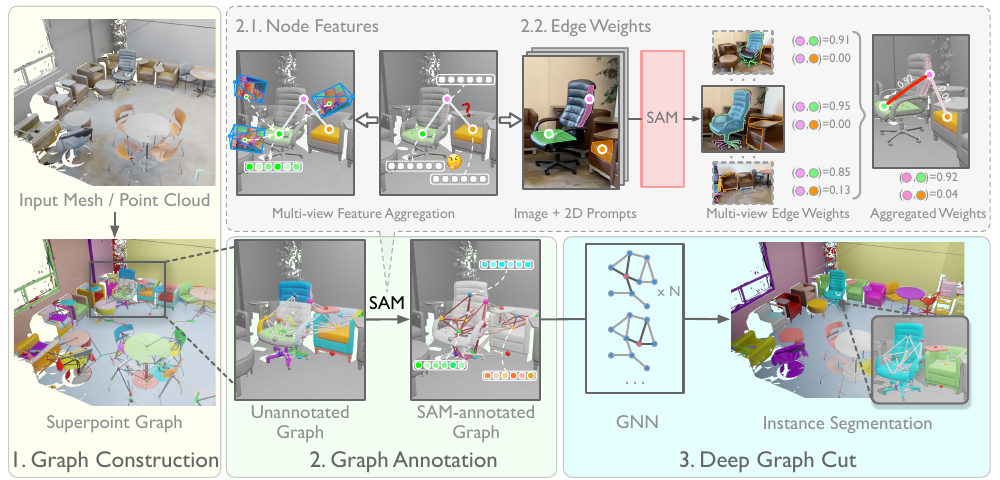}
\caption{
\textbf{Overview of our pipeline.} Our proposed 3D instance segmentation pipeline consists of three main parts.
\textbf{1.} We over-segment the input mesh / point cloud into superpoints and construct the structure of the superpoint graph based on adjacency (\cref{sec:superpoint}).
\textbf{2.} We utilize the prompt mechanism of SAM~\cite{sam} to annotate the nodes and edges of the graph (\cref{sec:sam}).
The node features are aggregated from multi-view SAM backbone features corresponding to each superpoint. The edge weights are calculated based on the intersection ratio between the multi-view SAM masks corresponding to each pair of superpoints that constitute an edge.
\textbf{3.} We use a graph neural network to further process the SAM-annotated graph and perform graph cut based on the calculated edge affinity scores to obtain the instance segmentation results (\cref{sec:graph_cut}).
}
\label{fig:pipeline}
\end{figure*}

Given 3D geometry and calibrated multi-view images of a scene, our goal is to obtain its 3D instance segmentation.
In this paper, we propose a novel segmentation framework, as illustrated in \cref{fig:pipeline}.
We first perform over-segmentation on the 3D geometry to generate a set of superpoints, reformulating the task into a graph cut problem (\cref{sec:superpoint}).
Then, \cref{sec:sam} describes how to leverage SAM to construct node features and edge weights of the superpoint graph.
In \cref{sec:graph_cut}, we introduce a graph neural network for 3D segmentation, which is trained with pseudo labels generated by 2D segmentation predictions.

\subsection{Building the superpoint graph}
\label{sec:superpoint}

In an indoor or outdoor scene, not only can we easily acquire multi-view images, but we can also obtain the scene's geometry (either in point clouds or mesh form) through depth cameras/laser scanners (for indoor scenes) or LiDAR (for large-scale outdoor scenes).
With the geometry of the scene available, we can proceed to pre-segment the scene based on traditional methods to obtain a set of superpoints.
For mesh, we apply the method in \cite{scannet}, which calculates the similarity between mesh vertices based on their normal directions and then conducts the graph cut algorithm \cite{felzenszwalb2004efficient}.
For point clouds, we employ the method in \cite{guinard2017weakly}, which first computes a local geometric feature vector (dimensionality and verticality) for each point, then performs Potts energy segmentation \cite{dann2012pottics}.

We can formulate the scene's instance segmentation task as a graph cut problem by employing superpoints. 
Specifically, we first represent the scene as a graph $G=(V,E)$, where $V$ denotes the set of superpoints in the scene and $E$ denotes the adjacency relationships between these superpoints.
Two superpoints are considered as adjacent if their distance is within a predefined threshold.
By employing superpoints, we can simplify the segmentation task in two ways. Firstly, the number of superpoints is substantially lower than the number of points in the original point cloud. Secondly, superpoints serve as a 3D proxy, enabling us to utilize multi-view image information and SAM's prompt mechanism to determine the connection between regions in 3D space.

\subsection{Constructing edge weights and node features}

To accomplish segmentation of the 3D scene, our primary task is to determine whether two superpoints on each edge should be merged.
To this end, we leverage multi-view image information and employ SAM to annotate the graph so that we can apply graph cut for segmentation.
Specifically, we utilize the prompt mechanism of SAM to annotate the edges and use SAM encoder features to annotate the nodes.

\PAR{Prompt mechanism of SAM.}
\label{sec:sam}
Unlike previous 2D instance segmentation methods that take an image as input and output its segmentation map, SAM (Segment Anything Model) \cite{sam} operates by taking an image and a prompt as inputs and producing corresponding segmentation result.
A typical prompt could be one or several 2D points.
The image and the prompt are fed into an image encoder and a prompt encoder separately.
Subsequently, a transformer-based decoder computes the cross-attention between prompt features and image features to generate the mask.
Specifically, SAM can output multiple valid masks with associated confidence scores. In our experiments, we tend to prefer masks with a larger area because they are more likely to represent an entire object.
We only resort to selecting masks with a relatively smaller area when the confidence of the larger mask is low (please refer to the supplementary materials for detailed implementation).

\begin{figure}[t]
\centering
\includegraphics[width=0.5\linewidth]{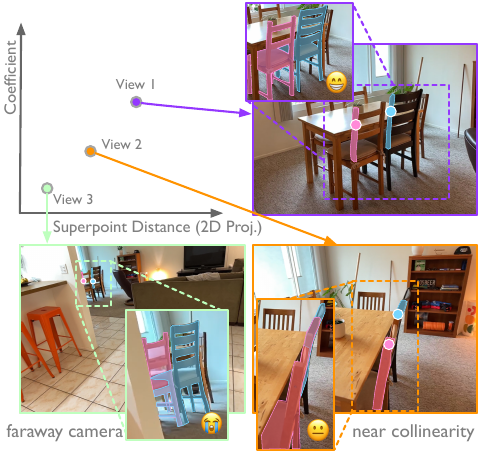}
\caption{
\textbf{Relationship of coefficient and 2D superpoints distance.}
For two superpoints, their distance in 2D images will be farther under near and frontal views than faraway or collinear views.
We assume that SAM achieves better performance on near and frontal views.
Thus, we consider the 2D distance as a factor in calculating the coefficient of each view.
}
\label{fig:distance}
\end{figure}

\PAR{Edge weights.}
The prompt mechanism of SAM introduces flexibility to calculate the edge weights between two 3D superpoints.
Part 2.2 of \cref{fig:pipeline} presents an illustration of computing the weight.
Specifically, we first select a view where both superpoints are visible (if such a view does not exist, we regard the weight to be zero).
Then, the two superpoints are projected onto the image space and $k$ points are uniformly sampled in each projection to serve as prompts for running SAM, thereby obtaining a mask for each superpoint.
For masks corresponding to the two superpoints, we examine their intersection situation.
Specifically, if two superpoints of an edge correspond to masks $A$ and $B$, then we calculate the edge weight as $w=\max(\frac{A\bigcap B}{A},\frac{A\bigcap B}{B})$, where $A \bigcap B$ denotes the intersection of $A$ and $B$.

Taking into account that two superpoints may be co-visible in multiple views, we calculate edge weights estimation across all these views and take their weighted average: $w=\sum_{i}c_i w_i$, where $w_i$ is the edge weight estimation of view $i$ and $c_i$ is the corresponding coefficient. $c_i$ is computed based on the following two factors: first, we consider the confidence of the two masks predicted by the SAM network; then, we consider the distance between the projections of the two superpoints in that view, as illustrated in \cref{fig:distance}.
Specifically, for a view, we obtain its score by multiplying the 2D distance of two superpoints with the confidence of two corresponding masks.
Then, we perform L1 normalization to multi-view scores to obtain coefficient $c_i$ for each view.

\PAR{Node features.} In addition to edge weights, we also annotate node features with SAM.
Specifically, for each node in the graph model, we identify all views that observe the corresponding superpoint. 
Within the projection range of that superpoint in each view, we randomly sample several points, interpolate to obtain features extracted by the SAM encoder and average the features obtained from all views to represent the attributes of the node.
Part 2.1 of \cref{fig:pipeline} presents an example of obtaining node features.

\subsection{Deep graph cut}
\label{sec:graph_cut}

We first propose a simple segmentation method that is not based on neural networks. Specifically, based on the superpoint graph constructed in \cref{sec:superpoint} and the edge weights of the graph calculated in \cref{sec:sam}, we use the edge weights to determine whether two superpoints are connected, then employ a method based on the union-find algorithm \cite{union-find} to merge all connected superpoints, thereby achieving 3D segmentation of the scene.

To improve the robustness of segmentation, we feed the graph model into a graph neural network (GNN) before segmentation. The GNN has a certain receptive field and can utilize the information of the surrounding nodes and edges and can predict edge affinity scores which can be more reliable than original edge weights.
To process the input graph, we design a GNN that consists of graph convolutional layers and fully connected layers.
The graph model is first passed through graph convolutional layers to extract features for each vertex. 
Then, we concatenate features of two vertices with the corresponding edge weight computed in Sec.~\ref{sec:sam}, which are fed into fully connected layers to predict the affinity between two vertices.
After that, we apply the same segmentation method based on the predicted affinity scores.

\PAR{Pseudo labels generation.}
We propose a strategy for training the GNN without 3D ground truth annotation.
To supervise the network, we first generate pseudo labels based on a 2D segmentation model.
For pseudo-labels, the most ideal case would be to obtain the correct affinity for all edges, but this is unrealistic. In fact, while we require a high degree of accuracy for pseudo-labels, the completeness of these labels is a lesser priority.
For this purpose, we use a 2D segmentation network, CropFormer \cite{cropformer}, for this task.
We first ran CropFormer on all views to obtain the instance segmentation results.
For each pair of co-visible superpoints in every view, we record whether these superpoints are within the same mask.
If they are co-visible in at least $n$ views and their records are consistent across all these $n$ views, then we treat the pair as a pseudo-label.
For example, if two superpoints are within the same mask in all co-visible views, they are treated as a positive sample in the pseudo-labels, vice versa.

The reason for choosing CropFormer is based on our empirical observation that it tends to yield relatively more complete and accurate masks for common object categories.
For example, CropFormer consistently segments a chair entirely, whereas SAM sometimes segments parts of it, such as a single chair leg.
Although CropFormer has its advantages, we opt for SAM in the previous graph construction stage due to the following considerations:
SAM's unique prompt mechanism can use superpoints to control the granularity of segmentation to some extent. Moreover, SAM's design, which allows for predicting multiple masks from a single prompt, makes it more adept at handling uncommon objects. Consequently, we chose SAM for constructing the entire graph, while using CropFormer to generate relatively incomplete but high-quality pseudo-labels.

\PAR{Training of the GNN.}
We employ pseudo-labels generated by CropFormer as supervision. For the edges included in these pseudo-labels, we compare the affinity scores of these edges predicted by GNN with the labels to calculate binary cross-entropy loss, which is defined as
\begin{equation}
    L_\text{BCE}=s_\text{ps}\log(s)+(1-s_\text{ps})\log(1-s),
\end{equation}
where $s$ is the predicted affinity score and $s_\text{ps}$ is the pseudo-label of corresponding edge.
Since pseudo-labels are sparse, we observed that direct training in this manner tends to limit the network's capabilities.
To improve the accuracy of the graph network's predictions, we introduce an additional regularization for edges not included in the pseudo-labels during training.
This regularization aims to ensure the predicted affinity score $s$, and the corresponding edge weight predicted by SAM, $w_\text{SAM}$, to be as consistent as possible.
This loss is defined as 
\begin{equation}
    L_\text{reg}=|w_\text{SAM}-0.5| L_1(s, w_\text{SAM}).
\end{equation}
With this design, the closer $w_\text{SAM}$ is to 0 or 1, the greater the penalty for inconsistency between $s$ and $w_\text{SAM}$. The final loss is defined as $L=L_\text{BCE}+L_\text{reg}$.

\section{Implementation details}
\label{sec:detail}

\PAR{Graph construction.} When sampling points within the projection of a superpoint as prompts for SAM, we uniformly sample $k=5$ points within the projected mask. Considering that there might be slight inaccuracies in the camera poses, the points too close to the boundaries of projection masks could potentially fall outside the object. Therefore, we take care to avoid sampling these points during this process.

\PAR{Training of GNN.} We implement the GNN with PyTorch \cite{pytorch} and PyG \cite{pyg}. When generating pseudo labels with CropFormer, we only consider the superpoint pairs that are co-visible in at least $n=10$ views with all consistent records. We train the GNN on ScanNet200 with pseudo-labels for 200 epochs, which takes about 20 minutes on an NVIDIA A6000 GPU.

\section{Experiments}

\begin{figure*}[t]
\centering
\includegraphics[width=\linewidth]{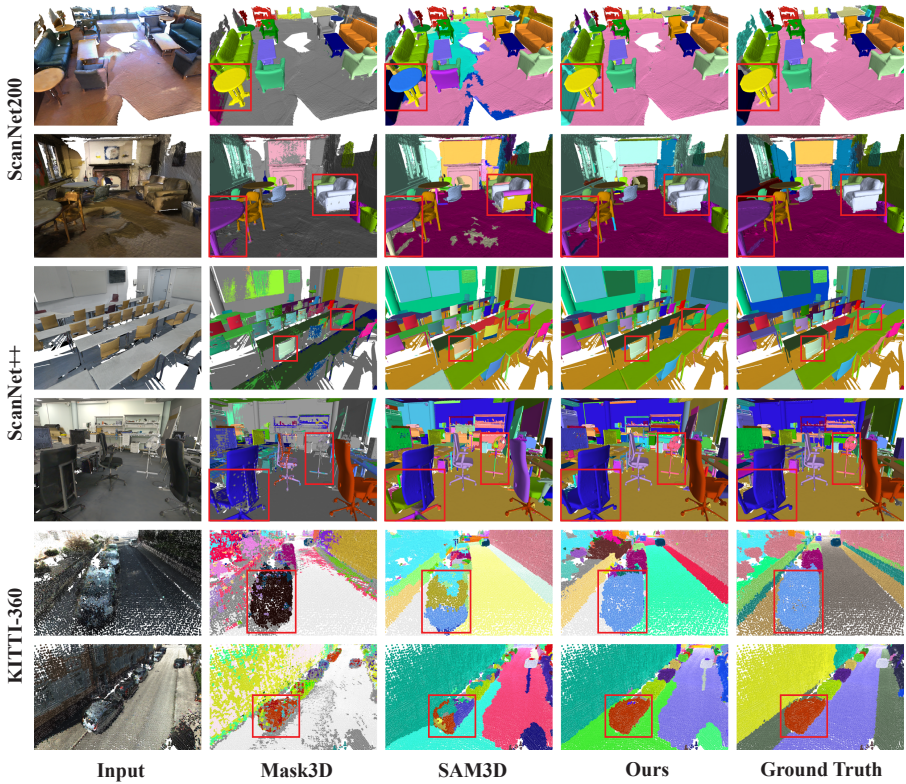}
\caption{\textbf{3D segmentation results on ScanNet200, ScanNet++ and KITTI-360 datasets.} Please zoom in for details.
Compared to Mask3D, our method exhibits significantly better generalization on ScanNet++ and KITTI-360 datasets. Moreover, in comparison to SAM3D, our approach can segment objects in the scene more completely and accurately.
We observed that Panoptic Lifting struggles to extract satisfactory geometry, so we leave the qualitative comparison with it to the supplementary material. 
}
\label{fig:comparison}
\end{figure*}
    
\begin{table*}[t]
\centering
\caption{\textbf{Quantitative results of 3D segmentation on ScanNet200, ScanNet++ and KITTI-360 datasets.} We report the AP scores averaged on all test scenes. Our method significantly outperforms the baseline methods on all datasets.}
\begin{tabular}{lcccccccccccc}
\toprule
&& \multicolumn{3}{c}{ScanNet200} && \multicolumn{3}{c}{ScanNet++} && \multicolumn{3}{c}{KITTI-360} \\
\cmidrule{3-5} \cmidrule{7-9} \cmidrule{11-13}
&& mAP & AP$_{50}$ & AP$_{25}$ && mAP & AP$_{50}$ & AP$_{25}$ && mAP & AP$_{50}$ & AP$_{25}$  \\
\midrule

Felzenswalb          &&  4.8 &  9.8 & 27.5 &&  8.8 & 16.9 & 36.1 && -    & -    & -    \\
Guinard              &&  2.9 &  8.2 & 33.1 &&  4.3 & 10.6 & 32.3 &&  9.3 & 18.9 & 39.6 \\
SAM3D (w/o ensemble) && 12.1 & 28.6 & 54.1 &&  3.0 &  7.9 & 22.3 &&  4.6 & 10.6 & 26.0 \\
SAM3D (w/ ensemble)  && 20.9 & 34.8 & 51.4 &&  9.3 & 16.6 & 29.5 && 13.0 & 24.2 & 41.1 \\
Ours + NCuts         && 15.7 & 31.7 & 59.0 && 10.1 & 18.6 & 34.7 && 19.5 & 30.2 & 45.0 \\
Ours + DBSCAN        && 10.3 & 18.6 & 27.8 && 10.5 & 17.2 & 25.0 && 20.5 & 31.4 & 42.1 \\
Ours (w/o GNN)       && 19.7 & 37.7 & 61.6 && 13.7 & 25.2 & 43.0 && 22.6 & 36.2 & 48.5 \\
Ours (w/ GNN)        && \textbf{22.1} & \textbf{41.7} & \textbf{62.8} && \textbf{15.3} & \textbf{27.2} & \textbf{44.3} && \textbf{23.8} & \textbf{37.2} & \textbf{49.1} \\
\bottomrule
\end{tabular}
\label{tab:comparison}
\end{table*}

\subsection{Datasets, metrics and baselines}
\PAR{Datasets.} We perform the experiments on ScanNet200 \cite{scannet200}, ScanNet++ \cite{scannet++} and KITTI-360 \cite{kitti360}.
ScanNet200 is built on ScanNet \cite{scannet}, which is a large-scale RGB-D dataset that contains 1613 indoor scenes with geometry acquired by BundleFusion \cite{bundlefusion} and images captured by iPad Air2.
ScanNet++ contains 280 indoor scenes with high-fidelity geometry acquired by the Faro Focus Premium laser scanner as well as high-resolution RGB images captured by iPhone 13 Pro and a DSLR camera with a fisheye lens.
KITTI-360 is a large outdoor dataset with 300 suburban scenes, which consists of 320k images and 100k laser scans obtained through a mobile platform in a driving distance of 73.7 km.
All of these datasets are annotated with ground truth camera poses and instance-level semantic segmentations.
The GNN in our method is trained on 1201 training scenes of ScanNet200 with our generated pseudo-labels, we then evaluate our method on 312 validation scenes of ScanNet200, 50 of ScanNet++ and 61 scenes of KITTI-360, according to the official split of each dataset.

\PAR{Metrics.} We evaluate our segmentation performance with the widely-used Average Precision (AP) score. We follow the standard defined in \cite{faster-rcnn,scannet,scannet200} for evaluation, calculating AP with thresholds of $50\%$ and $25\%$ (denoted as AP$_{50}$ and AP$_{25}$, respectively) as well as AP averaged with IoU thresholds from $50\%$ to $95\%$ with a step size of $5\%$ (mAP).
Since our method as well as most of baseline methods are class-agnostic, we do not consider semantic class label in evaluation, which follows the setting of \cite{rozenberszki2023unscene3d}.
Additionally, we exclude the predicted instances in unannotated regions for all methods to facilitate a fairer comparison.

\PAR{Baselines.} We compare our method with the following baselines: (1) Traditional segmentation methods: \cite{felzenszwalb2004efficient,guinard2017weakly}, which only use geometric information to perform segmentation. (2) 2D-to-3D lifting method: SAM3D \cite{sam3d} and Panoptic Lifting \cite{panoptic}. We report the results of SAM3D with and without the ensemble process. Since Panoptic Lifting is based on NeRF \cite{nerf} and requires hours for per-scene optimization, we only report corresponding qualitative analyses. (3) Point cloud segmentation method: Mask3D \cite{mask3d}, we use their official pretrained models. (4) We segment our SAM annotated graph (without GNN) with traditional spectral clustering methods: Normalized Cuts \cite{ncuts} and DBSCAN \cite{dbscan}, as well as the graph cut used in our method.

\subsection{Comparisons with the state-of-the-art methods}
We evaluate 3D segmentation metrics on ScanNet200, ScanNet++ and KITTI-360 datasets. Averaged quantitative results are shown in \cref{tab:comparison}. We also provide qualitative results in \cref{fig:comparison}.
By analyzing quantitative and qualitative results, we found that our method significantly outperforms state-of-the-art unsupervised methods.

Panoptic Lifting can achieve reasonably good segmentation results in simpler scenes, but its performance deteriorates in more complex environments with a large number of objects. SAM3D can handle complex scenes, however, its segmentation often results in both large structures, like floors, and smaller objects, like chairs, being divided into multiple segments. In contrast, our method is able to produce more accurate and complete segmentation results.
In the experiments of our method combined with Normalized Cuts and DBSCAN, we found that carefully tuning hyperparameters can yield relatively good result for a single scene. However, each scene varies significantly in scale and the number of objects, leading to considerable differences in the optimal hyperparameters. When we apply uniform hyperparameters across all dataset, the averaged metrics are not ideal.
\begin{table}[t]
\centering
\setlength{\tabcolsep}{4pt}
\caption{\textbf{Comparison with Mask3D.} While Mask3D shows better performance than ours on ScanNet200, it cannot generalize well to ScanNet++ and KITTI-360.}
\begin{tabular}{lllcccccc}
\toprule

&  & Training Data & mAP & AP$_\text{50}$ & AP$_\text{25}$ \\
\midrule
\multirow{4}{*}{ScanNet200}
& Mask3D         & GT of ScanNet200        & \textbf{53.3} & \textbf{71.9} & \textbf{81.6} \\
& Mask3D         & GT of ScanNetV2         & 45.1 & 62.6 & 70.5 \\
& Ours (w/o GNN) & -                       & 19.7 & 37.7 & 61.6 \\
& Ours (w/ GNN)  & Pseudo-GT of ScanNet200 & 22.1 & 41.7 & 62.8 \\
\midrule
\multirow{4}{*}{ScanNet++}
& Mask3D         & GT of ScanNet200        &  4.6 & 10.5 & 22.9 \\
& Mask3D         & GT of ScanNetV2         &  3.7 &  7.9 & 15.6 \\
& Ours (w/o GNN) & -                       & 13.7 & 25.2 & 43.0 \\
& Ours (w/ GNN)  & Pseudo-GT of ScanNet200 & \textbf{15.3} & \textbf{27.2} & \textbf{44.3} \\
\midrule
\multirow{4}{*}{KITTI-360}
& Mask3D         & GT of ScanNet200        &  0.2 &  0.9 &  7.0 \\
& Mask3D         & GT of ScanNetV2         &  0.3 &  1.0 &  8.0 \\
& Ours (w/o GNN) & -                       & 22.6 & 36.2 & 48.5 \\
& Ours (w/ GNN)  & Pseudo-GT of ScanNet200 & \textbf{23.8} & \textbf{37.2} & \textbf{49.1} \\
\bottomrule

\end{tabular}
\label{tab:comparison_mask3d}
\end{table}

\PAR{Comparisons with supervised method.} In addition to unsupervised baselines, we also compare our method with Mask3D, which is the state-of-the-art supervised learning method. We evaluate the class-agnostic AP scores, and the results are shown in \cref{tab:comparison_mask3d}. When evaluation, we use their official pretrained models.
Specifically, we first evaluate their model trained on the ScanNet200 training set, which shows very good results on the ScanNet200 validation set. However, when applied to the ScanNet++ dataset, there is a significant performance drop. This is because ScanNet++, although also indoor scene data, has a different data collection method from ScanNet200, indicating that the Mask3D method is quite sensitive to differences in aspects such as the point cloud collection method. When applied to KITTI-360, the performance becomes extremely poor, as KITTI-360 is an outdoor scene, which is quite different from ScanNet200 so that Mask3D cannot generalize well.
Then we evaluate Mask3D model trained on the ScanNetV2 training set. Compared to ScanNet200, ScanNetV2 lacks annotations for some categories of objects, resulting in decreased performance on the ScanNet200 validation set. This indicates that Mask3D has limited ability to generalize to different categories of objects even in the same type of scenes, and it also suffers significant performance declines on ScanNet++ and KITTI-360.

In contrast, our method (without GNN) achieves good results on all three datasets without any training, although our method performs worse on ScanNet200 than Mask3D, which is because Mask3D is trained with GT annotations provided by ScanNet200, allowing it to learn strong priors of this dataset. Our method shows a clear advantage on the ScanNet++ and KITTI-360 datasets. After using the GNN trained with pseudo-labels on ScanNet200, our method not only improves on the ScanNet200 validation set but also shows improvement on ScanNet++ and KITTI-360. This indicates that the GNN module in our method helps with segmentation and learns general prior for segmentation agnostic to the training dataset, thus has good generalization capabilities to data with large differences.

\subsection{Ablation studies}

To analyze the effectiveness of each module and design in our method, as well as the impact of hyper-parameters on performance, we conduct ablation studies on ScanNet200.

\begin{table}[t]
\centering
\setlength{\tabcolsep}{7pt}
\caption{\textbf{Ablation studies of graph cut implementation.}}
\begin{tabular}{lccc}
\toprule
&  mAP & AP$_{50}$ & AP$_{25}$ \\
\midrule
w/o GNN                   & 19.7 & 37.7 & 61.6 \\
w/o node features         & 19.4 & 37.5 & 61.3 \\
w/o edge weights          & 10.1 & 21.5 & 42.2 \\
w/o regularization loss   & 19.9 & 38.1 & 61.4 \\
our full method           & \textbf{22.1} & \textbf{41.7} & \textbf{62.8} \\
\bottomrule
\end{tabular}
\label{tab:ablation}
\end{table}
\begin{table}[t]
\centering
\begin{minipage}[t]{0.49\textwidth}
    
\centering
\setlength{\tabcolsep}{8pt}
\captionof{table}{\textbf{Ablation studies of $k$.}}
\begin{tabular}{lccc}
\toprule
$k$ &  mAP & AP$_{50}$ & AP$_{25}$ \\
\midrule
1  & 20.3 & 37.7 & 60.7 \\
3  & 21.2 & 38.8 & 61.2 \\
5  & \textbf{22.1} & \textbf{41.7} & \textbf{62.8} \\
7  & 22.0 & 40.1 & 62.1 \\
9  & 21.9 & 39.9 & 62.3 \\

\bottomrule
\end{tabular}
\label{tab:ablation-k}

\end{minipage}
\begin{minipage}[t]{0.49\textwidth}
    
\centering
\setlength{\tabcolsep}{8pt}
\captionof{table}{\textbf{Ablation studies of $n$.}}
\begin{tabular}{lccc}
\toprule
$n$ &  mAP & AP$_{50}$ & AP$_{25}$ \\
\midrule
1  & 20.3 & 38.7 & 61.2 \\
5  & 21.7 & 40.9 & \textbf{63.6} \\
10 & \textbf{22.1} & \textbf{41.7} & 62.8 \\
20 & 20.4 & 39.1 & 62.6 \\
30 & 20.2 & 39.3 & 61.9 \\

\bottomrule
\end{tabular}
\label{tab:ablation-n}

\end{minipage}
\end{table}

\PAR{Ablation studies of graph cut implementation.}
We evaluate with five configurations:
(1) Apply graph cut directly on edge weights of the SAM annotated graph (without GNN refinement).
(2) Annotate the graph without node features. (3) Annotate the graph without edge weights. (4) Train GNN without regularization loss. (5) Our full method. We report the quantitative results in \cref{tab:ablation}.
The results indicate that the absence of a GNN leads to a decline in segmentation performance.
Node features have a certain impact on the predictive performance of the GNN, while edge weights have a very significant impact, almost playing a dominant role. Additionally, we found that regularization loss is also important.

\PAR{Ablation studies of number of prompt points.}
We conduct ablation studies on the number of prompt points $k$ sampled in each superpoint projection, the results are shown in \cref{tab:ablation-k}.
In our experiment, we chose $k=5$, and the ablation results suggest that our method is not very sensitive to the value of $k$, but it is affected to some extent. The results are relatively poor when $k=1$, as sampling only one point within a projection mask leads to significant information loss, especially when the projection mask area is large. When $k$ is larger, the performance slightly deteriorates, which may be due to SAM not being suitable for accepting too many points as a prompt.

\PAR{Ablation studies of minimum number of co-visible views.}
We conduct ablation studies on the minimum number of co-visible views $n$ required for generating pseudo labels, the results are shown in \cref{tab:ablation-n}.
In our experiment, we chose $n=10$, and the ablation results indicate that our method is robust to the choice of $n$, but our performance is relatively poorer when $n$ is either too small or too large.
This is because when $n$ is small, a large number of unreliable pseudo-labels are generated. When $n$ is large, although the quality of the pseudo-labels is high, they become very sparse. In both cases, this leads to suboptimal results.

\PAR{Ablation studies of pseudo-label generation.}
Furthermore, we conduct experiments to analyze the advantages of using CropFormer to generate pseudo-labels compared to SAM. For the labels generated by CropFormer, we also use SAM to predict a label. The accuracy scores of both methods are evaluated using the ground truth annotations provided by ScanNet200. As indicated in the \cref{tab:analyse_cropformer}, CropFormer achieves higher accuracy.

\section{Conclusion}

In this paper, we introduced a novel 3D segmentation method with SAM guided graph cut. The key idea is to pre-segment 3D scenes into superpoints, and then utilize the prompt mechanism of SAM to assess the affinity scores between superpoints. We propose a GNN based graph cut method to achieve robust segmentation, which is trained with pseudo-labels generated by a 2D segmentation network. Experiments showed that the proposed method is able to achieve accurate segmentation results and can generalize well to different datasets.

\begin{table}[t]
\centering
\setlength{\tabcolsep}{5pt}
\caption{\textbf{Analyses of the choice for pseudo-labels generation.}}
\begin{tabular}{lcccc}
\toprule
& Accuracy & Precision & Recall & F1-Score \\
\midrule
SAM        & 0.892 & 0.934 & 0.888 & 0.911 \\
CropFormer & \textbf{0.912} & \textbf{0.934} & \textbf{0.923} & \textbf{0.929} \\
\bottomrule
\end{tabular}
\label{tab:analyse_cropformer}
\end{table}

\PAR{Discussion.} Our method not only requires geometric data (mesh/point cloud) but also needs multi-view images as input, which to some extent limits its application scenarios.
Moreover, we perform segmentation based on merging superpoints. When an object is part of a superpoint, we are unable to segment it out. This situation occurs occasionally, for example, a poster adhered to a wall.
To address this, a more sophisticated pre-segmentation model, which consider not only geometric information but also semantics, should be designed. One viable approach is to also use the guidance from SAM or other visual models during the pre-segmentation stage. We leave it to future works.

\PAR{Acknowledgement.} The authors would like to acknowledge the support from NSFC (No. 62322207), Ant Group and Information Technology Center and State Key Lab of CAD\&CG, Zhejiang University.

\bibliographystyle{splncs04}
\bibliography{main}

\clearpage
\appendix
    
\section{Superpoint generation}
For the mesh of each scene on ScanNet and ScanNet++, we use the segmentator \cite{segmentator} provided by ScanNet, which adopts the algorithm described in \cite{felzenszwalb2004efficient} based on the mesh's normals. Specifically, for ScanNet, we directly use the over-segmentation provided in their dataset, which is obtained by running the segmentator with \texttt{KThresh} set to $0.01$ and \texttt{segMinVerts} set to $20$. For ScanNet++, due to the higher point cloud density of ScanNet++, we adjust the parameters to \texttt{KThresh} as $0.2$ and \texttt{segMinVerts} as $500$ to run the segmentator. For KITTI-360, we use the unsupervised point cloud segmentation algorithm proposed in \cite{guinard2017weakly}, which first computes geometric features for each point from its 3D position and color values and then obtains the partitioned superpoints by minimizing the global energy function, and we set the regularization strength $\rho$ to $0.1$.

\section{Points sampling in projection mask}
To sample $k=5$ points uniformly and not too far from the boundary of the projection mask of each superpoint in each view, we first use the Euclidean Distance Transform implemented by OpenCV \cite{opencv} to compute the distance from each pixel within the mask to its boundary, creating a distance map. We then select the point with the maximum value in the distance map to ensure it is near the center. To prevent subsequent sampled points from being too close to this first point, we set the values in the distance map within a certain area around this point to zero. This process is iteratively repeated for sampling the remaining points.

\section{Multi-scale mask selection}

The 2D segmentation model SAM is designed to output masks at three different scales, each with a corresponding confidence score, to enable segmentation at different granularities. Please refer to their paper for detailed information. In our pipeline, we tend to choose masks with larger areas, as they are more likely to correspond to the segmentation of a complete object. However, when the confidence of a larger area segmentation is high, we will consider masks of smaller areas.
Specifically, if the confidence of the largest mask is higher than the others, or lower but within a margin of $0.05$, then we choose this mask. If this criterion is not met, we select from the remaining two masks: if the medium-sized mask has a higher confidence than the smallest mask, or its confidence is lower but within a margin of $0.05$, then we choose the medium-sized mask; otherwise, we choose the smallest mask.

\section{Structure of GNN}
The Graph Neural Network (GNN) in our method consists of a 5-layer Graph Convolutional Network (GCN) and a 3-layer Multi-Layer Perceptron (MLP). The GCN has an input channel size of 256, which corresponds to the channel size of the SAM features. It has a hidden layer width of 128 and an output channel size of 128. The MLP has an input channel size of 257, which includes the concatenated GCN features of two nodes and one edge weight. Its hidden layer width is 128, and it has an output channel size of 1, corresponding to the affinity score of an edge.

\section{Implementation details of graph cut}

When performing segmentation, for every two vertices, if their affinity is below a certain threshold, we consider them to be unconnected. 
Conversely, if their affinity is above this threshold, to further improve robustness, we identify paths of length 2 between these two vertices.
We then record the number of paths where both edges have high affinity scores and the number where one edge is high and the other low. 
If the ratio of the latter exceeds a predefined threshold, we regard the two vertices to be unconnected; otherwise, they are considered as connected. 
Once the connection status of each edge is determined using this method, we employ a union-find algorithm \cite{union-find} to merge all connected superpoints, resulting in the 3D segmentation of the scene.

\section{Details of evaluation protocol}
We evaluate the class-agnostic AP scores of all methods across all datasets, meaning that during the evaluation, we only consider the accuracy of the masks, without taking into account their semantic categories. This approach follows that of \cite{rozenberszki2023unscene3d}. Since our primary focus is on object instance segmentation, and the baseline Mask3D cannot segment the floors and walls, we exclude predictions of floor and wall regions (only the floor for KITTI-360) from evaluation for all methods to ensure a more fair comparison. Additionally, we also exclude segmentations predicted in other unlabeled regions.

\begin{figure*}[t]
\centering
\includegraphics[width=\linewidth]{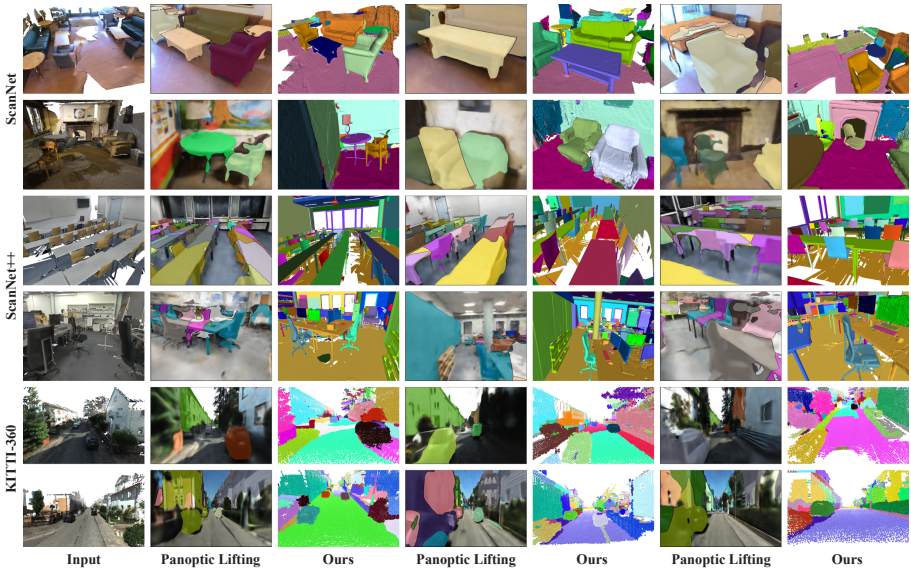}
\caption{\textbf{Comparison with Panoptic Lifting.}}
\label{fig:comparison_with_panoptic_lifting}
\end{figure*}

\section{Comparison with Panoptic Lifting}

We observed that Panoptic Lifting struggles to extract satisfactory geometry, so we render the results of Panoptic Lifting in several views and visualize our method in nearby views for comparison. We show the results in \cref{fig:comparison_with_panoptic_lifting}.

\section{Analyses of different graph cut method}

Based on the graph constructed using SAM, we tested segmenting the graph using normalized cuts, DBSCAN, and the direct graph partition method used in our approach, both with and without using the GNN (without means directly using edge weight). The comparison results are shown in the \cref{tab:ablation_graph_cut}. From the results, it's evident that the use of GNN generally improves most metrics for normalized cuts, while DBSCAN and the direct graph partition method show comprehensive improvements across all metrics. Furthermore, regardless of the use of GNN, the direct graph partition method consistently outperforms both normalized cuts and DBSCAN. Our analysis suggests that while normalized cuts and DBSCAN are adept at obtaining a rough segmentation for graphs with unreliable edge affinities, they are less capable of achieving finer segmentation results even when edge affinities are highly reliable.

\begin{table*}[t!]
\centering
\caption{\textbf{Ablation studies of different graph cut methods.}}
\begin{tabular}{lcccccccccccc}
\toprule
&& \multicolumn{3}{c}{ScanNet} && \multicolumn{3}{c}{ScanNet++} && \multicolumn{3}{c}{KITTI-360} \\
\cmidrule{3-5} \cmidrule{7-9} \cmidrule{11-13}
&& mAP & AP$_{50}$ & AP$_{25}$ && mAP & AP$_{50}$ & AP$_{25}$ && mAP & AP$_{50}$ & AP$_{25}$  \\
\midrule
NCuts            && 15.7 & 31.7 & 59.0 && 10.1 & 18.6 & 34.7 && 19.5 & 30.2 & 45.0 \\
DBSCAN           && 10.3 & 18.6 & 27.8 && 10.5 & 17.2 & 25.0 && 20.5 & 31.4 & 42.1 \\
Graph partition  && 19.7 & 37.7 & 61.6 && 13.7 & 25.2 & 43.0 && 22.6 & 36.2 & 48.5 \\
GNN + NCuts      && 18.0 & 35.0 & 59.4 && 11.3 & 20.1 & 35.2 && 18.1 & 27.7 & 40.7 \\
GNN + DBSCAN     && 11.0 & 19.6 & 29.2 && 10.7 & 17.5 & 25.9 && 21.1 & 32.2 & 43.0 \\
GNN + Graph partition  && \textbf{22.1} & \textbf{41.7} & \textbf{62.8} && \textbf{15.3} & \textbf{27.2} & \textbf{44.3} && \textbf{23.8} & \textbf{37.2} & \textbf{49.1} \\
\bottomrule
\end{tabular}
\label{tab:ablation_graph_cut}
\end{table*}
    
\section{Discussions of SAM guidance}

As shown in the ablation studies in our paper, both the node features and edge weights calculated based on SAM are effective for our method, with the edge weights being particularly crucial. To further analyze their effectiveness, we attempted to remove both and use PointNet++ to compute node features. Specifically, we utilized PointNet++ to extract features from the point cloud, averaging the features within a superpoint to serve as the node feature. We employed the same loss function as in our method and optimized the network parameters of both PointNet++ and the GNN simultaneously. We found that this approach resulted in very poor performance.

\section{Performance on small objects}

\begin{figure}[t]
  \centering
    \begin{minipage}{0.38\linewidth}
    {
        \includegraphics[width=1\linewidth]{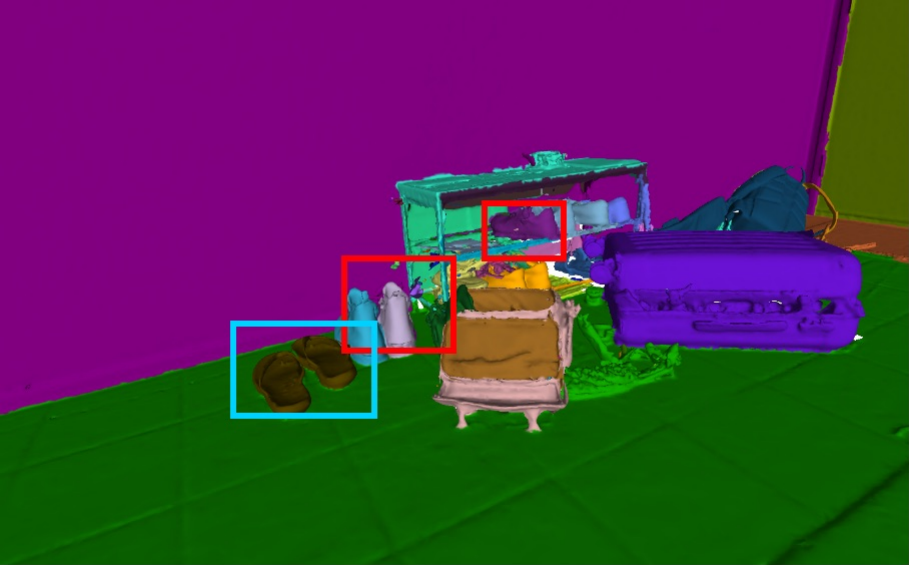}
        \subcaption{\label{fig:small1}}
    }
    \end{minipage}
    \begin{minipage}{0.38\linewidth}
    {
    	\includegraphics[width=1\linewidth]{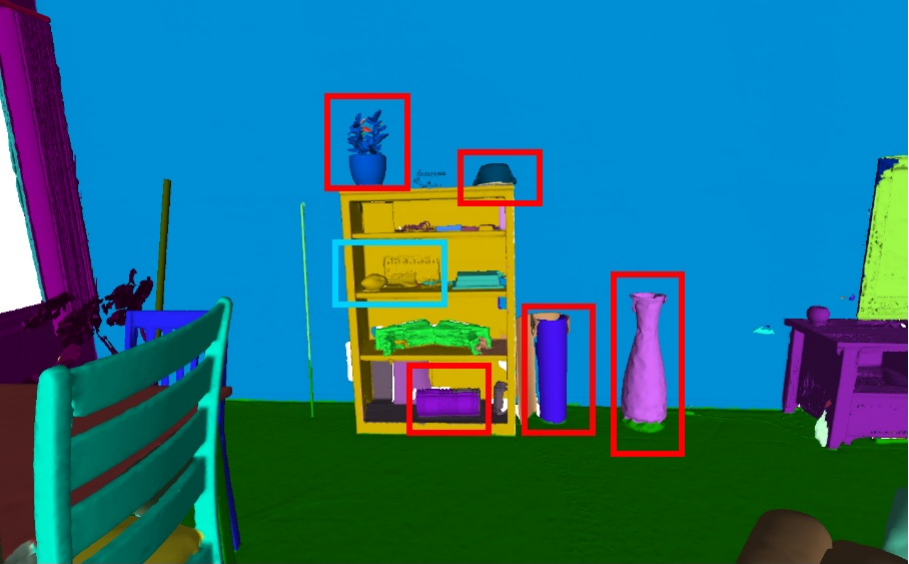}
        \subcaption{\label{fig:small2}}
    }
    \end{minipage}
  \caption{Segmentation performance on small objects.}
  \label{fig:small}
\end{figure}

We show segmentation results on small objects in \cref{fig:small}. Our method successfully segments a considerable number of small objects, such as several shoes in \cref{fig:small1} and some vases and small items on the shelf in \cref{fig:small2}, as framed in red, though some small objects are not separated from each other or their surroundings, as framed in blue.

\section{Failure cases}
\begin{figure}[t]
  \centering
  \begin{minipage}{0.38\linewidth}
    {
        \includegraphics[width=1\linewidth]{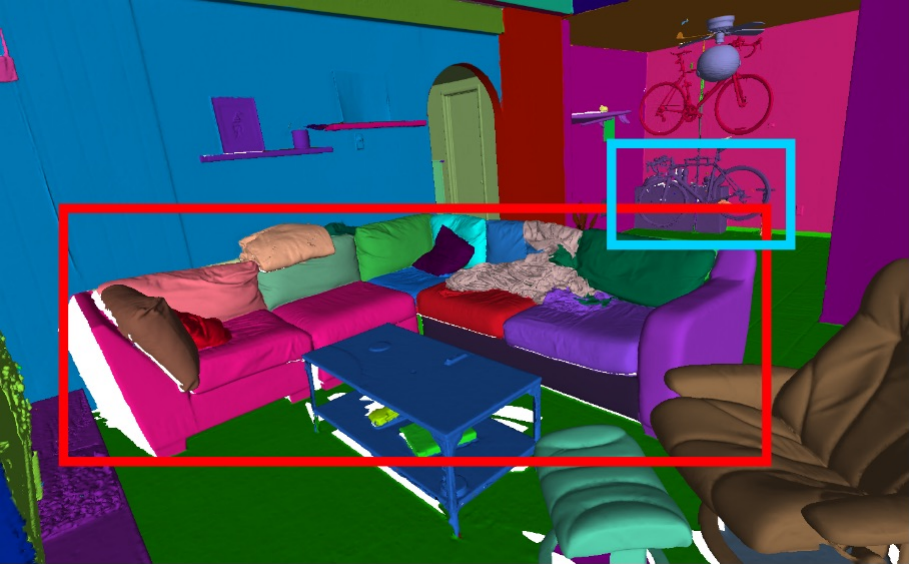}
    }\end{minipage}
    \begin{minipage}{0.38\linewidth}
    {
    	\includegraphics[width=1\linewidth]{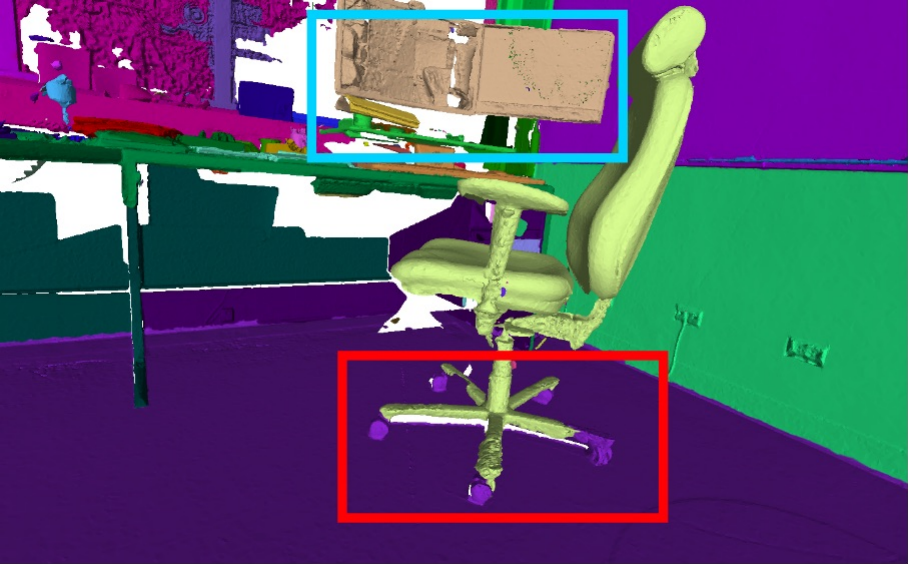}
    }\end{minipage}
  \caption{Failure cases.}
  \label{fig:failure}
\end{figure}

To better understand the performance of our method, we show two typical failure cases (over-segmention and under-segmention) of our method in \cref{fig:failure}, framed in red and blue respectively.

\end{document}